\documentclass{vldb}
\usepackage{graphicx}
\usepackage{balance}  

\usepackage{epstopdf}
\usepackage{subfigure}
\usepackage{multirow}
\usepackage{diagbox}
\usepackage{booktabs} 
\usepackage{url}
\usepackage{amsmath}
\usepackage{color}
\usepackage{soul}
\usepackage{comment}
\usepackage{CJK}

\newtheorem{definition}{Definition}

\vldbTitle{TitAnt: Online Real-time Transaction Fraud Detection in Ant Financial}
\vldbAuthors{Shaosheng Cao, Xinxing Yang, Cen Chen, Jun Zhou, Xiaolong Li, and Yuan Qi}
\vldbDOI{https://doi.org/10.14778/xxxxxxx.xxxxxxx}
\vldbVolume{12}
\vldbNumber{xxx}
\vldbYear{2019}

\begin{document}


\title{TitAnt: Online Real-time Transaction Fraud Detection \\in Ant Financial}



%
%
%
%

\numberofauthors{6} 

\author{
%
%
\alignauthor  Shaosheng Cao\\
       \affaddr{AI Department (Hangzhou)}\\
       \affaddr{Ant Financial Services Group}\\
       \affaddr{556 Xixi Rd, Hangzhou, China}\\
       \email{shaosheng.css@antfin.com}
\alignauthor  XinXing Yang \\
       \affaddr{AI Department (Beijing)}\\
       \affaddr{Ant Financial Services Group}\\
       \affaddr{9F East Tower, WFC, 1 East 3rd Ring, Beijing, China}\\
       \email{xinxing.yangxx@antfin.com}
\alignauthor  Cen Chen \\
       \affaddr{AI Department (Singapore)}\\
       \affaddr{Ant Financial Services Group}\\
       \affaddr{1 Raffles Place, Singapore}\\
       \email{chencen.cc@antfin.com}
\and  
\alignauthor  Jun Zhou \\
       \affaddr{AI Department (Beijing)}\\
       \affaddr{Ant Financial Services Group}\\
       \affaddr{9F East Tower, WFC, 1 East 3rd Ring, Beijing, China}\\
       \email{jun.zhoujun@antfin.com}
\alignauthor  Xiaolong Li \\
       \affaddr{AI Department (Seattle)}\\
       \affaddr{Ant Financial Services Group}\\
       \affaddr{500 108th A. NE Bellevue, Washington 98004, USA}\\
       \email{xl.li@antfin.com}
\alignauthor Yuan Qi\\
       \affaddr{AI Department (Hangzhou)}\\
       \affaddr{Ant Financial Services Group}\\
       \affaddr{556 Xixi Rd, Hangzhou, China}\\
       \email{yuan.qi@antfin.com}
}



\maketitle

\begin{abstract}

With the explosive growth of e-commerce and the booming of e-payment, detecting online transaction fraud in real time has become increasingly important to Fintech business.
To tackle this problem, we introduce the \texttt{TitAnt}, a transaction fraud detection system deployed in Ant Financial, 
one of the largest Fintech companies in the world. 
The system is able to predict online real-time transaction fraud in mere milliseconds. We present the problem definition, feature extraction, detection methods, implementation and deployment of the system,  as well as empirical effectiveness. Extensive experiments have been conducted on large real-world transaction data to show the effectiveness and the efficiency of the proposed system. 

\end{abstract}

\section{Introduction}

Fraud, such as phone fraud, insurance fraud and credit card fraud, causes severe problems for government and business. However, detecting such a fraud has always been challenging. With the rapid development of the e-commerce and e-payment, the problem of online transaction fraud has become increasingly prominent. Compared with traditional areas, online transaction is facing a considerably larger volume of fund transfer.

According to the statistics \cite{report}, in the year of 2017, the number and the volume of online transaction reaches 48 billion and 2, 075 trillion yuan respectively only in China. Ant Financial\footnote{https://en.wikipedia.org/wiki/Ant\_Financial}, also known as Alipay, accounts for about 58\% of China's third-part online payment transactions \cite{alipay}. Specifically, on 2017's Double Eleven Shopping Festival\footnote{https://en.wikipedia.org/wiki/Singles\%27\_Day} (similar to Black Friday Day in the US), a single day's transaction shot up to US\$25 billion \cite{mitrev,double11}.  With such transaction volume, it becomes thus of great significance to detect and prevent online transaction fraud.



To collect and analyze such a magnitude of transaction data, it requires a robust database component for offline storage and management. Furthermore, a large-scale distributed computing component for running algorithms is also necessary. To satisfy the low latency requirements for online serving, online prediction with efficient data accessing is of great significance. 
Meanwhile, feature extraction and detection methods are equally important.

Rule-based methods have been extensively studied over the years \cite{phua2010comprehensive} for fraud detection problem. However, fraud patterns change rapidly over time, greatly deteriorating the effectiveness of rules summarized by expert experience. Subsequently, 
many data mining based methods have been investigated. For example, supervised learning methods, are proposed recently \cite{ngai2011application,sagar2016online}. 
However, transaction data usually exhibit two kinds of characteristics: 1) the labels are unbalanced, i.e., the majority of transactions are not fraudulent but normal, and 2) compared with analyzing individual transaction records, aggregated data often provides much richer information to identify fraud patterns. 

To cope with the first characteristic, several unsupervised learning and anomaly detection methods are introduced \cite{casas2016poster, liu2008isolation}, 
however label information can hardly be utilized. On the other hand,  some existing data aggregation strategies are also applied for detecting fraud \cite{whitrow2009transaction,jha2012employing}, nevertheless, most of the previous approaches can hardly capture the complex fraud patterns of the online transactions. It is this paper's topic to investigate how to deal with these two characteristics with our methods.

In this paper, we present a real-world task in FinTech and introduce our TitAnt\footnote{It indicates the combination of Titan and Ant, where the Titans are giant deities with incredible strength in Greek mythology and the Ant is the name of the company by the meaning of ``greatness comes from micro things''. } system, which is actively detecting fraudulent transactions. Our contributions are summarized as follows:

\begin{itemize}
\item  We carefully analyze the task and some discoveries are excavated. Based on our observations, new feature extraction approaches for transaction fraud detection are 
examined, which is capable of making  full use of the information from aggregated data.
\item We design and develop a real-world transaction fraud detection system which is able to train offline large-scale data in hours, and predict online real-time transaction fraud within only milliseconds.
\item We conduct extensive experiments on a large transaction record dataset to validate the effectiveness and efficiency of our system, including rule-based methods, anomaly detection approaches and classification models.

\end{itemize}

Our paper is organized as follows. Section 2 discusses related work of fraud detection. Section 3 presents the problem definition, feature extraction and detection methods. Section 4 describes the details of the implementation and deployment of our TitAnt system. Section 5 shows experimental results, followed by the conclusion in Section 6. 

\section{Related Work}
In this section, we investigate the related literature, including expert systems and rule-based approaches, supervised and unsupervised learning algorithms for fraud detection task, as well as recently proposed network representation learning models. 

\subsection{Rule-based Methods and Expert System}

 Quinlan \cite{quinlan1990learning} and Cohen \cite{cohen1995fast} introduce assertion statement of IF \{conditions\} and THEN \{a consequent\} to recognize fraud records at first. By distinguishing fraudulent and normal records,  Brause et al. \cite{brause1999neural} generalizes and weighs the association rules of detecting credit card fraud. Based on previous achievement, Baulier et al. \cite{baulier2000automated} identifies implicit fraudulent calls by generating decision variables, Rosset et al. \cite{rosset1999discovery} investigates a two-stage rule-based solution to detect telephone fraud, and Wheeler and Aitken \cite{wheeler2000multiple}  adopt case-based reasoning to analyze the hardest ones of misclassified cases. Expert system based methods, on the other hand, also have been well investigated. Major and Riedinger \cite{major2002efd} uses statistical knowledge to construct a five-layer system, Von Altrock \cite{von1996fuzzy}, Stefano and Gisella \cite{stefano2001insurance}, Pathak et al. \cite{pathak2005fuzzy} respectively design different fuzzy expert systems for a specific scene. Besides, Chiu and Tsai \cite{chiu2004web} proposes FPM algorithm to mine frequent patterns of credit card transactions. With the rapid evolution of fraud patterns, only hand-summarized rules or expert knowledge are not sufficient to satisfy today's online detection, the methods \cite{quinlan1986induction, kuhn2013applied} learning knowledgeable information from historical data is more worthwhile to investigate.

\subsection{Supervised Learning Models}

Hand \cite{hand1981discrimination} first uses a linear discriminative model to detect fraud, and later Foster and Stine \cite{foster2004variable} propose an improved least square regression with stepwise selection predicting. Bayesian approaches have been investigated, where Ezawa and Norton \cite{ezawa1996constructing} employ a four-stage Bayesian network model for telephone fraud and Viaene et al. \cite{viaene2004case} adopts AdaBoosted naive Bayes for insurance fraud. Otherwise, neural network based models are applied in fraud diagnosis \cite{ghosh1994credit, patidar2011credit, aleskerov1997cardwatch}. Subsequently,  Syeda et al. \cite{syeda2002parallel} develops a parallel system of fuzzy neural networks, Barse et al. \cite{barse2003synthesizing} leverages the memory-based neural network to capture temporal dependencies, and Maes et al. \cite{maes2002credit} combines Bayesian networks and neural networks for detecting credit card fraud. Also, Bhowmik  \cite{bhowmik2011detecting} applies Bayesian classification and decision trees in insurance fraud detection task. Besides, Kim et al. \cite{kim2003design} and Wang and Ma \cite{wang2012hybrid} utilize SVM-based ensemble strategy for detecting telecommunication subscription fraud and credit fraud. Besides, Halvaiee and Akbari \cite{halvaiee2014novel} and Jia-jie \cite{jia2012electronic} respectively investigate the effectiveness of the artificial immune system and particle swarm optimization algorithm in fraud detection.

\subsection{Unsupervised Learning and Aggregation Strategies}

Cox et al. \cite{cox1997brief} visualizes data with the information of color, position, size and etc. to help to detect fraud. Bolton et al. \cite{bolton2001unsupervised} introduces profiling method to detect credit card fraud, Burge and Shawe-Taylor \cite{burge2001unsupervised} uses a recurrent neural network to exploit temporal information of account behavior, and Cortes et al. \cite{cortes2003computational} explores graph mining algorithms such as link analysis. Later, Yamanishi et al. \cite{yamanishi2004line} detects the fraud from medical insurance data by recognizing statistical outliers. Aggregated data analysis is also investigated, in which Perlich and Provost \cite{perlich2003aggregation} propose a novel target-dependent aggregation method, Casas et al. \cite{casas2016poster} utilizes k-means to classify network security data, 
and Vadoodparast et al. \cite{vadoodparast2015fraudulent} combines the results of several different clustering methods. In addition,  Jha et al. \cite{jha2012employing} and Whitrow et al. \cite{whitrow2009transaction} detect credit card fraud employing transaction aggregation. Anomaly detection methods, such as isolation forest \cite{liu2008isolation}, sheds light on fraud detection tasks, since fraudulent transactions are undoubtedly regarded as abnormal cases.



\subsection{Network Representation Learning Models}
Recently, network representation learning, also known as graph embeddings, plays an increasingly important role in network analysis. Perozzi et al. proposes DeepWalk \cite{perozzi2014deepwalk}, which is superior to  traditional graph analysis approaches like Spectral Clustering \cite{tang2011leveraging}, Modularity \cite{tang2009relational}, and wvRN \cite{macskassy2003simple}. After that, many models are introduced, for example, LINE \cite{tang2015line}, GraRep \cite{cao2015grarep}, node2vec \cite{grover2016node2vec} and etc. Besides, Structure2Vec \cite{dai2016discriminative} is a state-of-the-art supervised fashion of generating embeddings. Although these models have been demonstrated to be effective on the public dataset, there does not exist a distributed version that is able to support real-world industrial-scale transaction records.



\begin{figure*}
  \centering
  \subfigure[Basic features are extracted from user profile, transfer environment and etc.]{\includegraphics[scale=0.6]{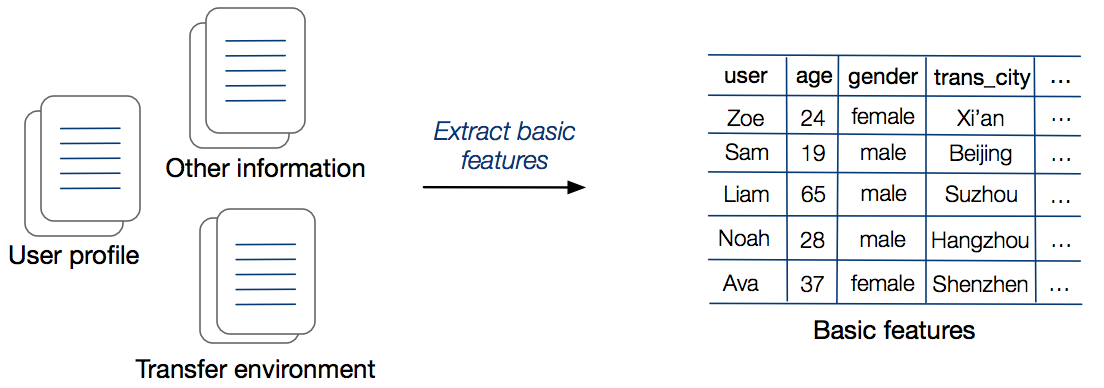}} \\
  \subfigure[User node embeddings are learned from historical transaction records.]{\includegraphics[scale=0.6]{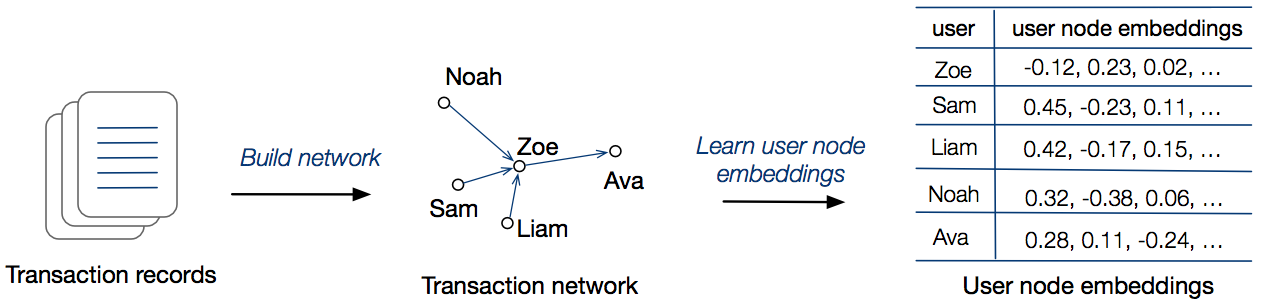}}
  \caption{An illustrated example of basic features extraction and user node embeddings generation.}
\label{user_emb}
\end{figure*}

\begin{figure} [htbp!]
  \centering
  \includegraphics[scale=0.2]{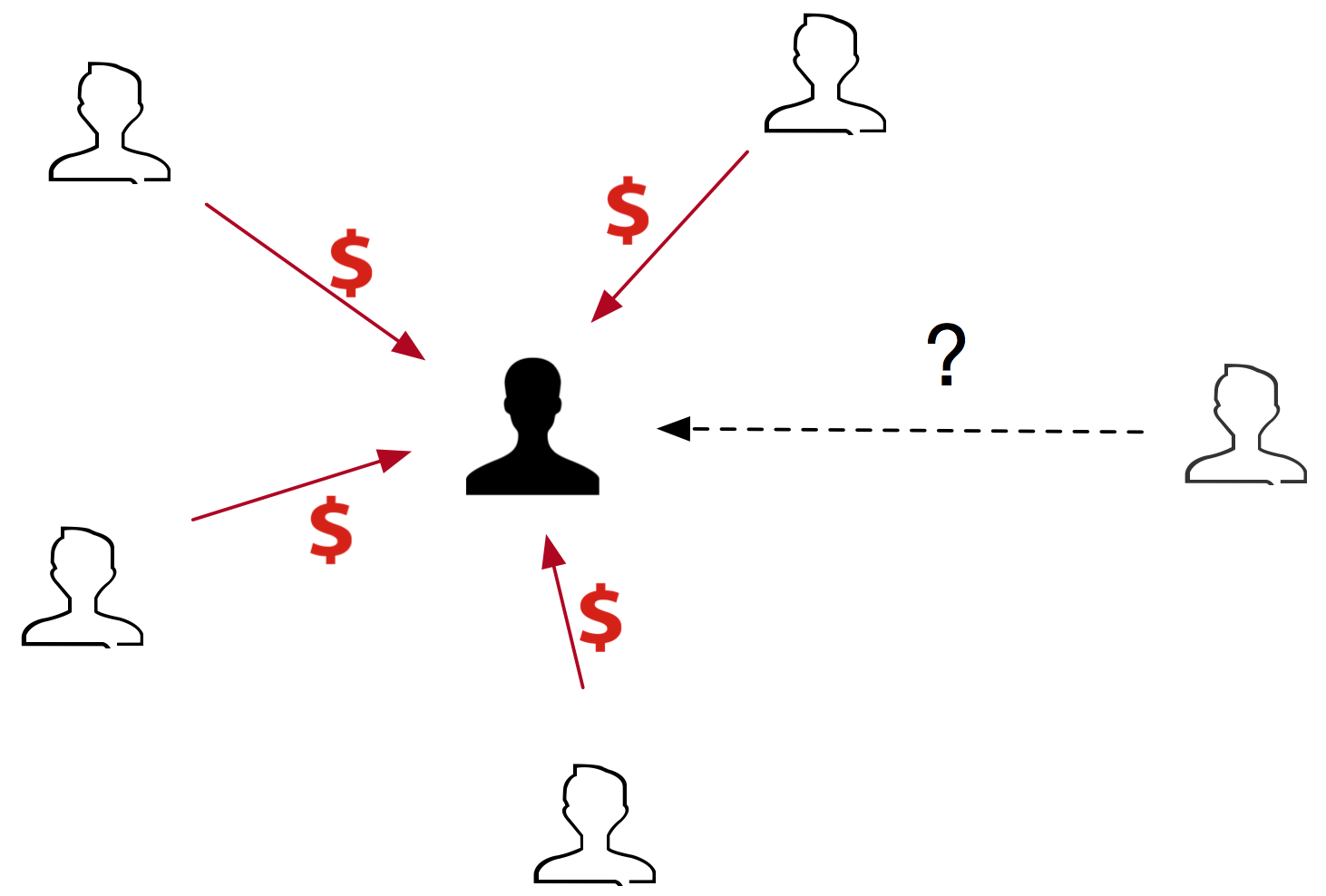}
  \caption{A simple case of aggregated data over the transaction network.}
\label{smplcs}
\end{figure}

\section{Problem Definition, Feature Extraction and Detection Methods} \label{method}

In this section, based on our analysis of the problem, feature extraction and detection methods are introduced. 



\subsection{Problem Definition}
In general, online transaction fraud can be categorized into two different types: explicit and implicit.  In an explicit case, a user is aware of the fraud afterwards. After a transaction is completed, the user could file a fraud report and upload the supporting proofs. Based on the transaction details, profiles and evidence, the authenticity of transaction fraud will be examined. If this user indeed suffers from a fraud, the fraudsters would be punished with punitive measures, such as action restrictions or account lockout, but it would be difficult to recover the losses according to the laws. This type is defined as an explicit fraud after an accident. 

In an implicit case, what we are concerned is to take proactive actions to prevent the potential event of fraudulent transactions, i.e., actively detecting online transaction fraud and taking immediate steps to prevent suspicious transactions. Contrary to explicit fraud, implicit one reveals less information and requires real-time prediction of the system. In this paper, we aim to tackle the implicit online real-time transaction fraud detection task and a formal problem definition is described as follows:

\begin{definition}
(Online Real-time Transaction Fraud Detection) Given historical transaction records with fraud labels, the task of \textit{online real-time transaction fraud detection} is to design a system to predict whether an online real-time transaction is a fraud or not.
\end{definition}

\begin{figure*} [htbp!]
  \centering
  \includegraphics[scale=0.5]{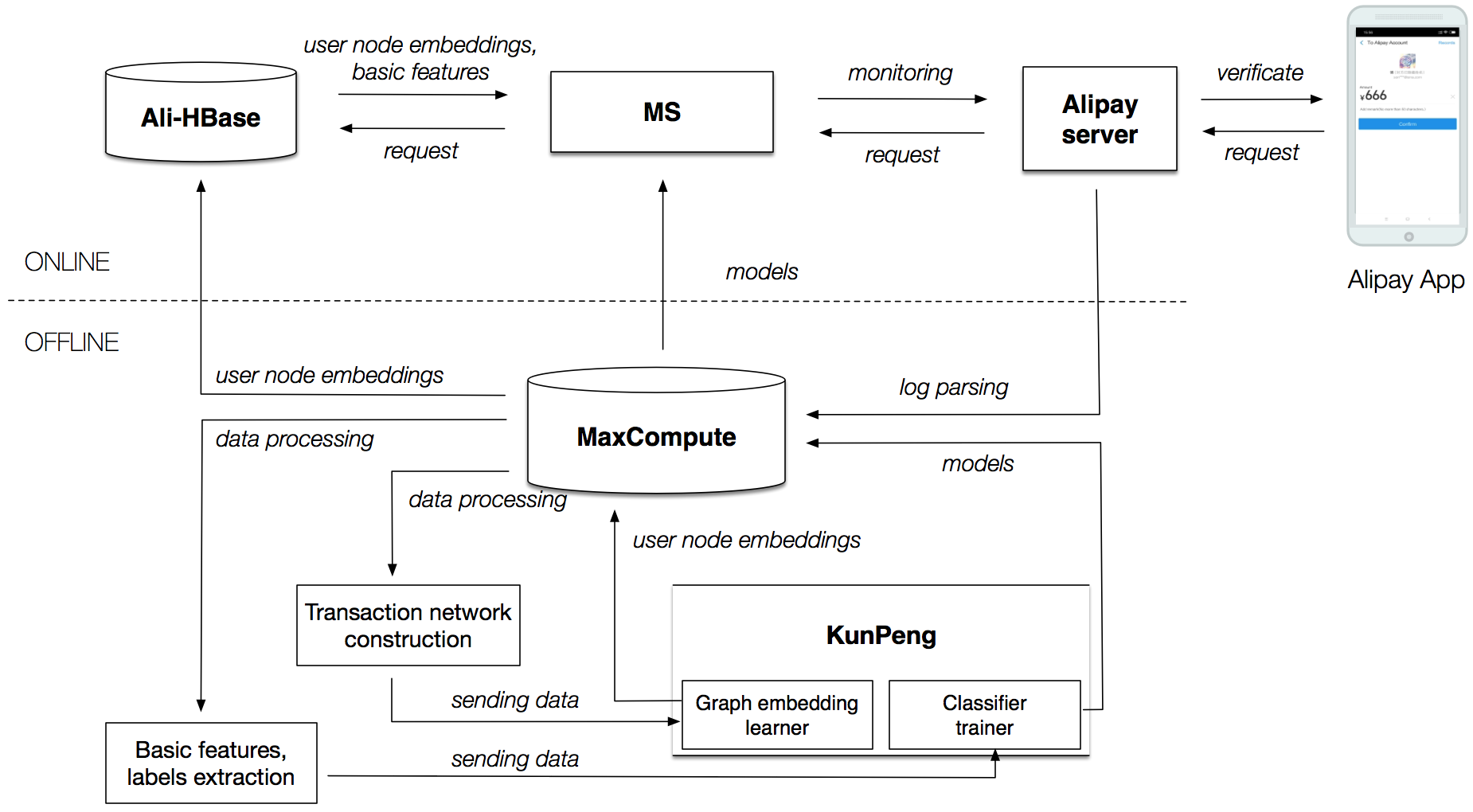}
  \caption{The architecture of TitAnt system.}
\label{titant_sys}
\end{figure*}

\subsection{Feature Extraction from Aggregated Data }

In order to discover transaction fraud, user profile and transfer contextual information are often of great importance. In particular, the fraudulent rates in some specific locations are always higher than other areas. 
Figure \ref{user_emb} (a) illustrates the basic user profile features extracted, such as age, gender, and transfer city (trans\_city)\footnote{trans\_city can be inferred from transfer IP address.}.


In addition, aggregated information on transaction records can provide much richer information.  Based on our investigation, approximately 70\% of the fraudsters have fraudulent behaviors more than once. It suggests that fraudsters tend to repeat their deceitful actions once successful. 
In Figure \ref{smplcs}, we give a simple example to demonstrate the value of aggregated data.
A directed edge reflects the transfer relationship from the corresponding transferor to the transferee. Directed red lines with a dollar sign indicate the fraudulent transactions, while a black user node stands for the fraudster. 
The on-going transaction, i.e., the dashed line with a question mark, is very likely to be a potentially implicit fraud.  Such gathering behaviors are often observed in the real cases and  manifest in more complex ways.

To extract useful information from the aggregated transaction data, a  transaction network is leveraged. Formally, we define the transaction network as follows:

\begin{definition}
(Transaction Network) A \textit{transaction network} is defined as $G=(V,E)$. $V=\{v_{1}, v_{2},\ldots, v_{n}\}$ is a collection of {\em nodes} with each node $v$ indicating a user while $E=\{e_{i,j}\}$ is a set of {\em edges} with each edge $e$ indicating the transfer relationship from a transferor to a transferee, both regarded as user nodes. 
\end{definition}

Based on historical records in a period of time, transaction network is built for analysis. Recall the simple case in Figure \ref{smplcs}, all the victims including the potential one have a same neighbor, i.e., the fraudster. It suggests they are 2-hop neighbors to each other. 
Therefore, the analysis of topological relationship is worthy of well studying in the transaction network. 
To capture topological relationship information,  Network Representation Learning (NRL) is a promising direction to be explored \cite{zhang2018network}.
Given a transaction network, NRL methods aim to learn a low dimensional representation matrix ${D}\in\mathbb{R}^{|V|\times{}d}$, whose $i$-th row ${D}_i$ is a $d$-dimensional vector representing the node $v_i$ in the transaction network.
In this way, the topological information can be captured by dense vectors, i.e., node embedding. 
Figure \ref{user_emb} (b) shows the procedures of generating user node embeddings. First, historical transaction records are collected to construct transaction network, and then user node embeddings are learned by NRL methods.

As most NRL implementations in the literature are limited to a single machine, we need to reimplement in a distributed learning framework, since huge amount of transaction records are being produced every day. Based on the insights that no one NRL method is the best in all cases \cite{goyal2017graph}, we select DeepWalk (\textbf{DW}) \cite{perozzi2014deepwalk} for its efficiency, effectiveness and simplicity.


Original DW utilizes random walk to generate short node sequences which transforms the topological information from the network into the sequences. Intuitively, the neighbors of one node will often occur in its contextual position in the linear node sequences. After the linear node sequences are generated, Skip-gram with negative sampling in word2vec \cite{mikolov2013distributed} is applied to generate user node embeddings finally. 

We also reimplement Structure2Vec (\textbf{S2V}) \cite{dai2016discriminative} as an alternative. Such supervised method can take full advantages of label information, but the learned user node embeddings are also affected by unbalanced labels. 
Meanwhile, unsupervised methods like DW do not require any labels, therefore, the topological information is extracted only from transaction network without being influenced by the imbalance of labels. 

\begin{figure*} [htbp!]
  \centering
  \includegraphics[scale=0.48]{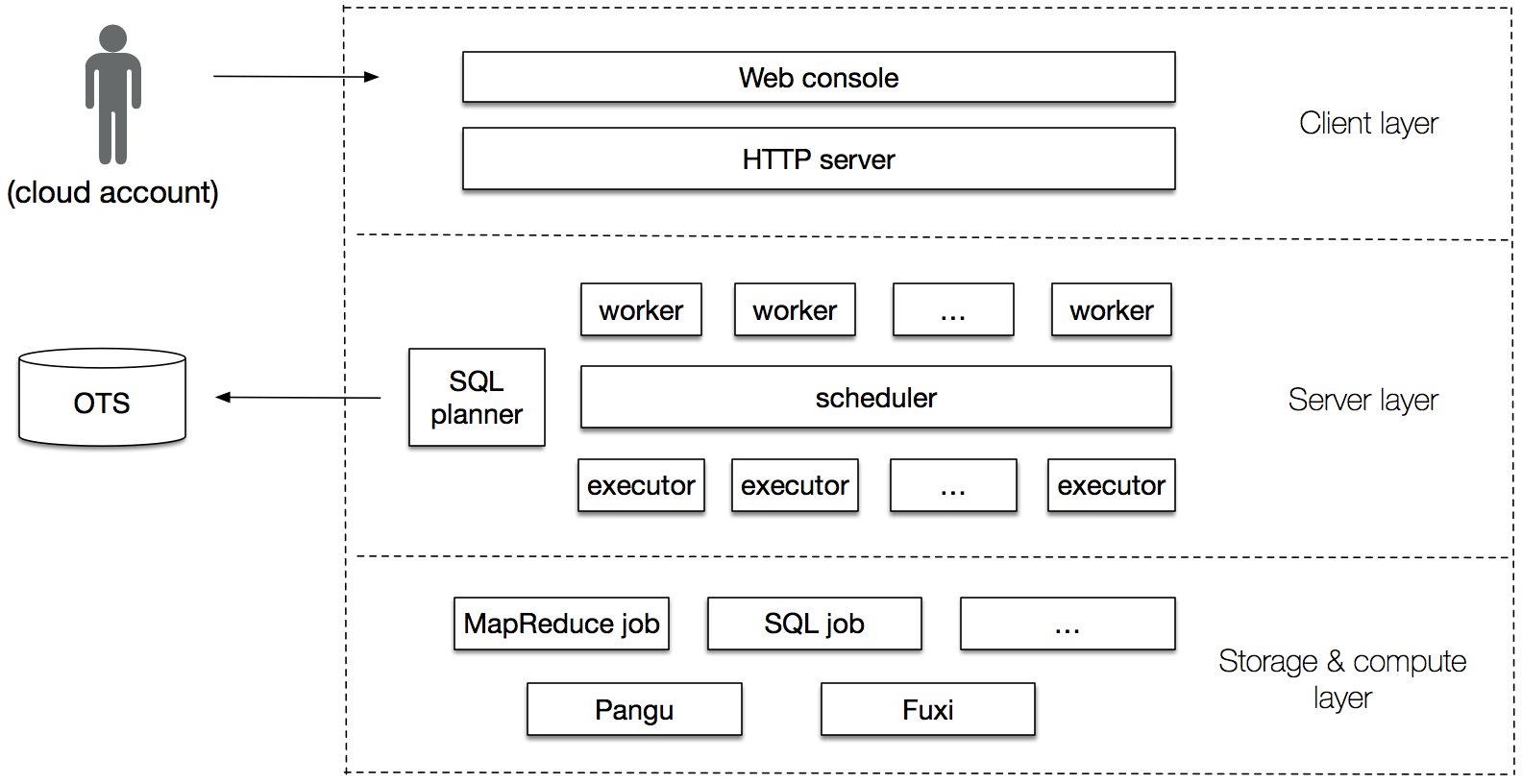}
  \caption{The architecture of MaxCompute.}
\label{maxcompute}
\end{figure*}

\subsection{Detection Methods}

As the problem of fraud detection is vital to a Fintech business, efforts have been spent for years, where about fifty features are carefully engineered. We call such features as basic features, which are also treated as rules or attributes. For each user, we generate user node embeddings, i.e., aggregated features, as additional information from the aggregated transaction records.
Basic features and aggregated features are then concatenated together. 
Labels are collected from user fraud reports, thus cannot be obtained in real-time. 

In order to precisely find out fraud, we extensively investigate and validate rule-based methods, anomaly detection  approaches and classification models.

Rule-based methods are 
widely used in many fraud detection applications. Iterative Dichotomiser 3 (\textbf{ID 3}) \cite{quinlan1986induction} is a traditional approach based on decision tree learning, whereas \textbf{C5.0} \cite{kuhn2013applied, c50} is revised version of C4.5 \cite{quinlan2014c4} to extract informative patterns from data with higher accuracy. In those methods, features are regards as rules and label information is utilized to do fine-tune. 

Isolation Forest (\textbf{IF}) \cite{liu2008isolation} is a classical anomaly detection approach widely used due to its effectiveness. We treat features as attributes and directly predict fraudulent transactions, since it does not require any label information. 
Intuitively, transaction fraud detection is similar to anomaly detection tasks, since the goal is to find out abnormal transactions, i.e., outliers that are more likely to be separated from most of the other data. 

One of the most popular classification models is Logistic Regression (\textbf{LR}) \cite{walker1967estimation}.
Although continuous features can be used in LR, better performance can be achieved after feature discretization in most cases. 
Compared with LR, non-linear models such as, Gradient Boosting Decision Tree (\textbf{GBDT}) \cite{friedman2001greedy,friedman2002stochastic,ahmed2013application} is able to achieve better performance in a variety of industrial tasks.
GBDT is a tree-based classification model, whose decision trees learn the decision boundary of the classification dataset, and gradient boosting combines several weak classifiers into a stronger one.

We will examine the effectiveness of the above detection methods in Section \ref{exp}. 


\begin{figure*} 
  \centering
  \includegraphics[scale=0.5]{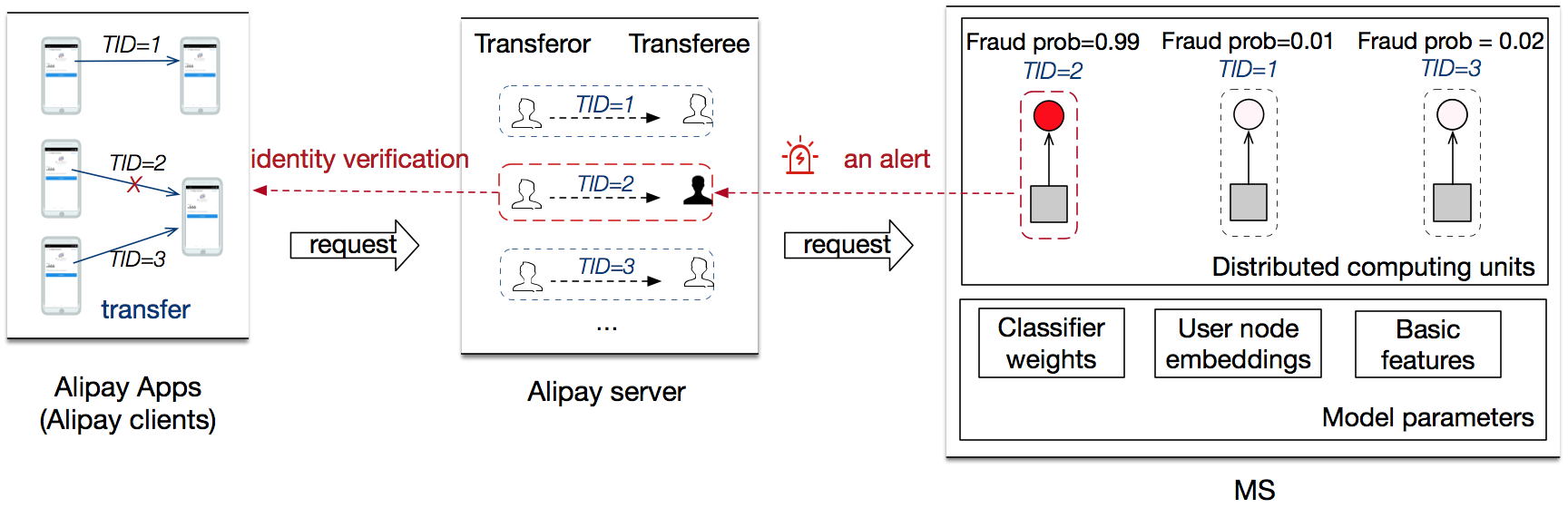}
  \caption{The architecture of the MS and its interactions with other components.}
\label{ms}
\end{figure*}

\section{TitAnt System implementation \\and deployment} \label{sys}

In this section, we show the details of the implementation and deployment of our TitAnt system. 

\subsection{The Framework of TitAnt System}

To guarantee timely response on fraud detection requests, low latency predictor, robust database storage platform, and distributed algorithms ought to be carefully designed. 
As illustrated in Figure \ref{titant_sys}, our system mainly has two parts, i.e., offline periodical training and online real-time prediction. In the offline training part, where models are trained on a fixed time basis, and model files are uploaded to online predictor for real-time transaction monitoring.  


Once users initiate transaction requests in Alipay\footnote{https://itunes.apple.com/us/app/alipay-simplify-your-life/id333206289?mt=8}, transaction logs will be periodically sent to MaxCompute\footnote{https://www.alibabacloud.com/product/maxcompute}  for offline computation. MaxCompute supports SQL and MapReduce for extracting basic features/labels and constructing transaction network. 
At the same time, KunPeng supports large-scale distributed NRL and classification model training\footnote{Only classification based detection methods are reimplemented in KunPeng for better performance, as reimplementation is time-consuming. Rule-based and anomaly detection methods are not distributed.}. The learned user node embeddings and classification models are stored in MaxCompute. 

Online prediction happens at Model Server (MS), where the model files are periodically updated. Once a transaction created by a user in Alipay APP, Alipay server immediately requests the Model server (MS). MS then gets the related data from Ali-HBase and makes real-time prediction. If the transaction is detected as fraud, the on-going transaction will be interrupted and transferor will be notified.
More details on each component will be elaborated in the following subsections. 

\subsection{MaxCompute}

MaxCompute, formerly known as Open Data Processing Service (ODPS), is a database storage and management platform. It has three logical layers: client layer, server layer and storage \& compute layer. As illustrated in Figure \ref{maxcompute}, developers can login with their cloud account and  submit jobs by web console in client layer, where HTTP server receives the command and send message to next layer.
Server layer contains workers, executors and scheduler to split jobs into subjobs for distribution. 
Also, heterogeneous jobs, such as mapreduce, SQL and etc., can be recognized and operated in the storage \& compute layer based on Pangu and Fuxi, where Pangu is a disk storage module and Fuxi is a resource scheduling module \cite{zhang2014fuxi}. 


When a SQL command is submitted by web console, the message is sent to the HTTP server, which requires the verification of cloud account information. If authentication passes, the job will be delivered to worker and the corresponding job instance will be sent to the scheduler. After that, scheduler registers the instance in Open Table Service (OTS) via SQL planner and its status is set as ``running'' simultaneously. OTS maintains the status of all the instances. Finally, scheduler adds the instance into the queue and corresponding instance ID will be generated.  

Subsequently, the scheduler will split the task of job instance into multiple subtasks, which are arranged into task pool in priority order. After that, scheduler keeps waiting for the available resource for computing. As soon as the resource conditions are satisfied, the subtasks are sent to an executor, which requests Fuxi to trigger computing resources in the compute layer.
When all the subtasks are finished, the executor updates the status of the instance as ``terminated'' in OTS. Finally, the results will be stored in Pangu.

\subsection{KunPeng}

\begin{figure} [htbp!]
  \centering
  \includegraphics[scale=0.27]{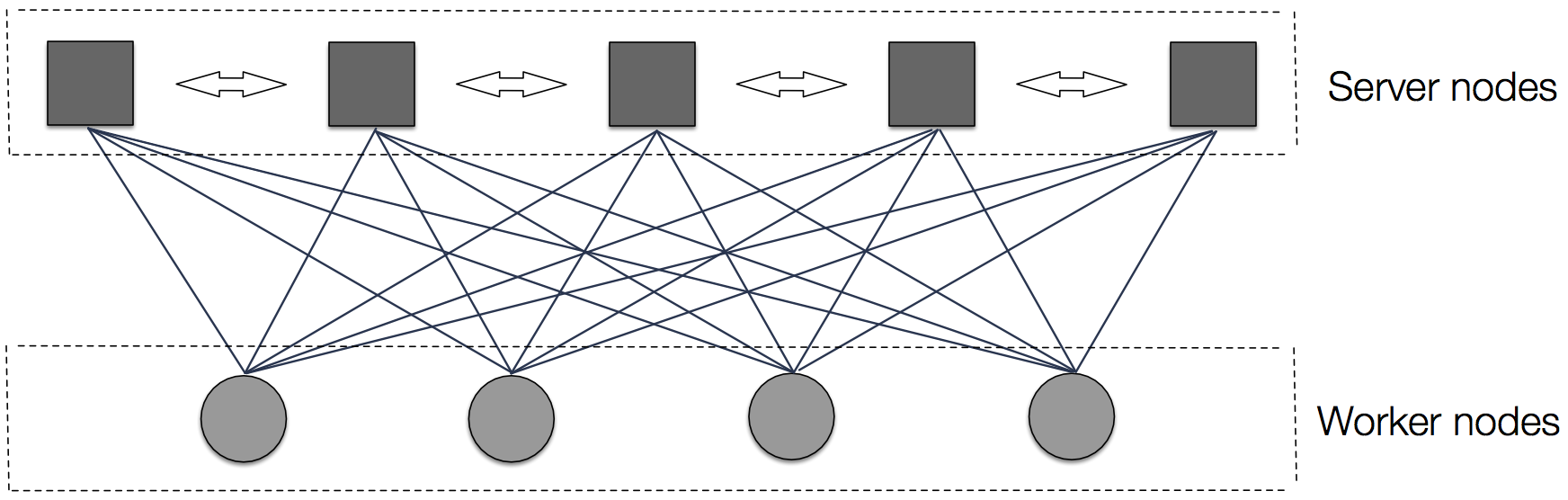}
  \caption{The system architecture of KunPeng.}
\label{kunpeng}
\end{figure}


As numerous transaction records wait for analysis every day, a distributed computing platform is an urgent need. Traditional frameworks, such as MPI \cite{gropp1999using}, do not support good failure tolerance. However, Parameter Server (PS) \cite{li2013parameter} supports a single point of failure, i.e., the failed instance can be restarted and recovered to the previous status automatically while other instances remain not affected. KunPeng system \cite{zhou2017kunpeng} is self-developed by the company based on PS framework , where various machine learning algorithms are running simultaneously.

KunPeng supports data parallelism and model parallelism. As illustrated in Figure \ref{kunpeng}, it consists of server nodes and worker nodes, where server nodes store the model parameters while worker nodes are responsible for training. Pull and Push operations are defined between server and worker nodes for data exchange. Besides, communication also happens among server nodes.


Based on KunPeng, we redesign NLR and classification algorithms, such as DW, S2V, LR, and GBDT. 
As an important part of DW, our reimplemented word2vec is involved in both worker and server nodes.
Worker nodes receive the node sequences by Random walk algorithm.  For every iteration, each worker first read a batch of sequence data and generate negative word list.  The embeddings are then pulled from server nodes and are updated by gradient descent. Subsequently, the updated embeddings are uploaded to server nodes. On the other hand, server nodes are responsible for communication with workers in order to exchange embedding data.  Server nodes first randomly initialize the embeddings and wait for the push requests from worker nodes. Once the push request is received, the corresponding embeddings are sent. After the update of each worker, server nodes pull the new embeddings and aggregate them by executing the model average operation.




\subsection{MS and Ali-HBase}

\begin{figure} [htbp!]
  \centering
  \includegraphics[scale=0.38]{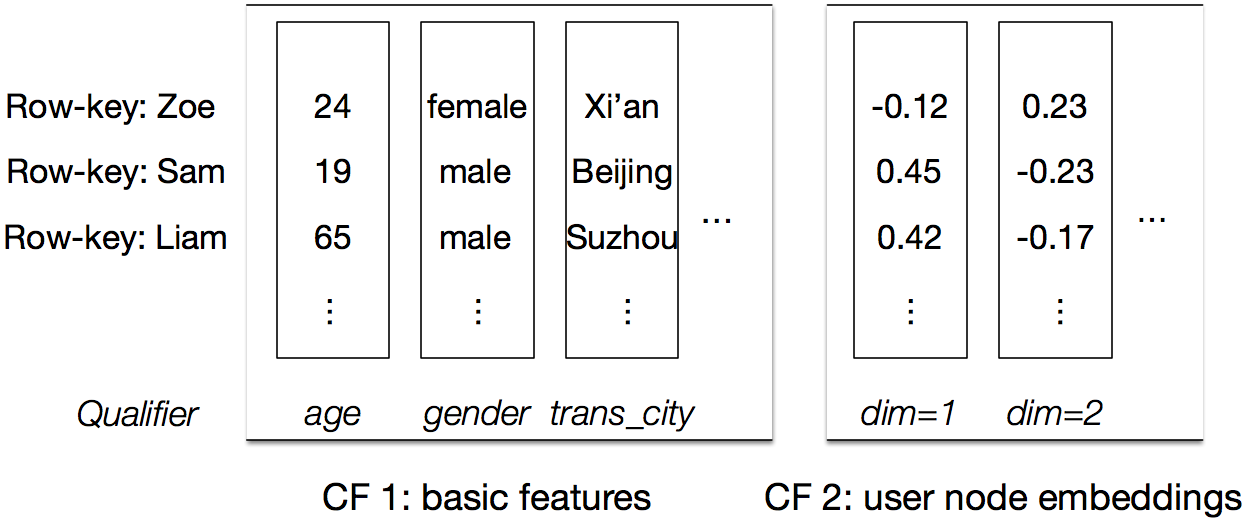}
  \caption{The architecture of Ali-HBase.}
\label{hbase}
\end{figure}

\begin{figure*} [htbp!]
  \centering
  \includegraphics[scale=0.46]{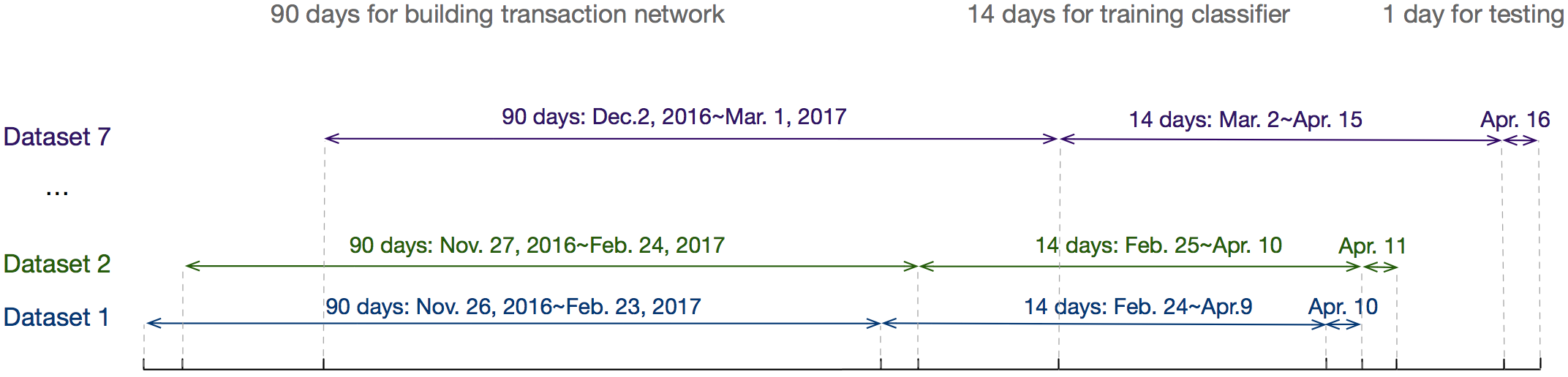}
  \caption{A graphical illustration of the datasets.}
\label{fig:time_seg}
\end{figure*}


Once offline training section ends, online real-time prediction works.
Figure~\ref{ms} shows an illustrative example of the whole real-time prediction process. When a user transfer money in Alipay App, the transfer request is sent to the Alipay server, followed by the MS for fraud monitoring. 
MS will access data from Ali-HBase for the latest version of user node embeddings and basic features. 
MS are distributed to satisfy low latency and high service load. 
As shown in Figure~\ref{ms}, the transaction TID=2 is probably a fraud with predicted fraud probability of 99\%,  thus MS sends an alert to the Alipay server, which will further interrupt the corresponding on-going transaction.


Ali-HBase is based on HBase Project\footnote{https://hbase.apache.org/}. HBase is first proposed as Bigtable \cite{chang2008bigtable}, a distributed, scalable and big data store, which is suitable for our real-time data accessing scenario. The inner data is organized in the form of Column Family (CF), where qualifier is used as a marker. 
As shown in Figure \ref{hbase}, the first CF is basic features where age, gender, and trans\_city are qualifiers. And the second is user node embeddings, where each dimension of value is the qualifier. 
In Figure~\ref{hbase}, users like Zoe, Sam and Liam are row-keys, to index the corresponding data.
Every time offline training is completed,  the data is uploaded to Ali-HBase by the version of date time.

\subsection{Discussion}


In this section, we discuss the implementation design, deployment issues and the construction of transaction network.




First, the system has strict serving requirements, i.e., tens of milliseconds at most for online detection including computation and communication costs. However, labels are usually delayed, as they are collected through user feedbacks, where online training is impractical. Thus, we adopt periodical offline training and real-time prediction in our system.

Second, in our system,  we only demonstrated the usefulness of user node embeddings learned from transaction network. One may ask what about other aggregated information, such as device and IP information? It is an interesting question to construct a heterogeneous network. We will explore this direction in future work.


\section{Experiments} \label{exp}

To empirically quantify the benefits of each component of our TitAnt system, 
we conduct experiments under different configurations. 

\begin{table*}
\centering
\caption{Performance under different eleven configurations.}
\vspace{6mm}
\label{table_res}
\resizebox{1.0\textwidth}{!}
{
\begin{tabular}{c|l|ccccccc} \hline 
Number&F1 Score&April 10&April 11&April 12&April 13&April 14&April 15&April 16 \\ \hline
1&Basic Features/Attributes+IF&10.30\%&10.38\%&11.62\%&11.21\%&10.82\%&11.00\%&13.30\% \\ \hline
2&Basic Features/Rules+ID3&42.08\%&44.72\%&41.21\%&44.25\%&42.33\%&41.94\%&47.69\% \\
3&Basic Features/Rules+C5.0&44.56\%&51.55\%&45.94\%&51.17\%&50.23\%&51.91\%&57.07\% \\ \hline
4&Basic Features+LR&53.08\%&58.47\%&55.72\%&60.13\%&56.87\%&52.52\%&64.38\% \\
5&Basic Features+GBDT&56.80\%&65.47\%&59.05\%&64.87\%&59.19\%&60.34\%&68.85\% \\ \hline
6&Basic Features+S2V+LR&55.21\%&62.08\%&60.78\%&64.11\%&61.04\%&55.83\%&68.86\% \\
7&Basic Features+S2V+GBDT&60.23\%&66.37\%&63.24\%&68.87\%&64.79\%&63.30\%&71.10\% \\ \hline
8&Basic Features+DW+LR&56.06\%&61.15\%&58.37\%&61.13\%&60.08\%&56.00\%&67.33\% \\
9&Basic Features+DW+GBDT&\textbf{61.43}\%&\textbf{66.87}\%&\textbf{64.11}\%&\textbf{69.93}\%&\textbf{65.10}\%&\textbf{64.00}\%&\textbf{71.84}\% \\ \hline
10&Basic Features+DW+S2V+LR&56.70\%&61.41\%&60.69\%&62.78\%&63.29\%&57.74\%&67.21\% \\
11&Basic Features+DW+S2V+GBDT&61.37\%&66.76\%&\textbf{64.11}\%&69.67\%&64.53\%&63.48\%&71.40\% \\ \hline
\end{tabular} }
\end{table*}

\subsection{Experimental Setup}

On this task, we adopt ``T+1'' mode to update the model, which means a model will be trained and deployed in an offline manner on a daily basis and will be used for prediction for the next day on a real-time basis. 
To demonstrate the effectiveness of our system, we have conducted several experiments and reported the performance of each day over a continuous week. In total, we have seven sets of data, where each one is sliced into three subsets: one for learning user node embeddings, another for training the classifier, and the last for testing. Specifically, we collect 90 days of transaction records to build the transaction network. 
The next 14 days of labeled records are treated as the training set  
and the last day of labeled records are used for the test set. 
 
For example in Dataset 1 (illustrated in Figure \ref{fig:time_seg}), transaction records of April 10, 2017 are chosen as the test set, 14 days' records prior to the test set are used as the training, and the earlier 90 days of records are employed to build the transaction network.
Different from other industrial scenes, such as e-commerce recommendation, online testing is hard to achieve since labels are not real-time obtained.

In our experiment, one of the goals is to investigate the effectiveness of basic features and the learned user node embeddings based on transaction network. More specifically, we compare both unsupervised DW and supervised S2V models on our task with unbalanced labels. For a fair comparison, the size of the learned embeddings is set to 32 and is concatenated with the basic features. In addition, for detection methods, we test the validity of rule-based ID3 and C5.0, anomaly detection based IF and classification based LR and GBDT.



For DW, we set the length of the random walk as 50, where each node is sampled as the first node of the sequences 100 times, i.e., the number of sampling is 100. It takes around 1.5 hours to learn the embeddings with approximate 8 million randomly selected transaction records with 20 machines equipped with 10 threads in our production environment. Aside from the transaction network, we also feed S2V with the fraud ground truth as the edge labels. Besides, there are 
a total of 52 basic features carefully extracted.

We set 100 trees for IF and raw basic features are fed as attributes. As rule-based ID3 and C5.0 cannot support continuous values well, we discretize the data into different bins \cite{kotsiantis2006discretization}.  
We impose L1 regularization and assign its weight as 0.1 for LR, and set 300 iterations as the stopping criteria.  For GBDT, we generate 400 trees with the depth of 3 to ensemble the results and use root mean square error as the objective. The subsampling rate of samples and features are set as 0.4 to prevent overfitting. 


\subsection{Empirical Results on Transaction Fraud Detection}

In this section, we empirically evaluate the effectiveness of our proposed system for the transaction fraud detection task. Eleven configurations are tested in Table \ref{table_res} from April 10 to April 16, where  F1 score is chosen as the evaluation metric. 
The best results are written in bold font for each day. 

First, we analyze the effectiveness of the learned user node embeddings. With the same classifier, it is obvious that introducing additional features from aggregated data can consistently improve the performance of the task. For example, on April 10,  the F1 score for ``Basic features+GBDT'' is 56.80\%. Adding the embeddings generated by S2V will improve the baseline by 3.4\% while adding the embeddings by DW will boost the performance by 4.6\%. The similar conclusion can be obtained for the rest of the days.


We can observe that using user node embeddings learned by DW leads to better results than S2V. Although supervised S2V utilizes extra label information, it also suffers from the issue of unbalanced labels. In this case, experimental results demonstrate that the benefits from the label information are weaker than the losses suffering from the label imbalance. Moreover, in the experiments, we further concatenate the different sources of the learned user node embeddings together from DW and S2V, but the performance is not improved compared with only DW is used. This suggests the topological information has already been well extracted by DW. 

 

\begin{figure} [htbp!]
  \centering
  \includegraphics[scale=0.46]{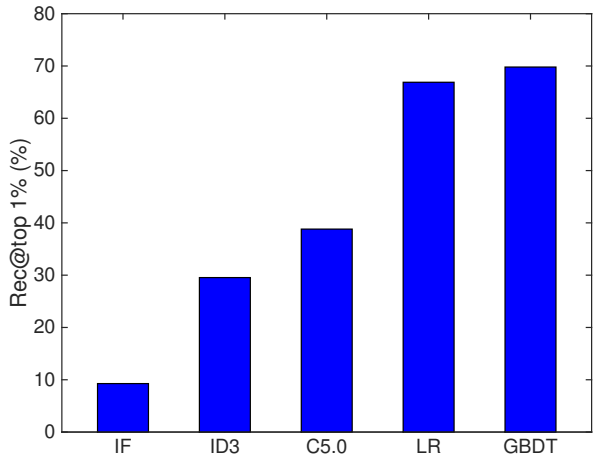}
  \caption{Recall scores for the top 1\% of the most suspicious frauds under different detection methods.}
\label{fig:perf_rec}
\end{figure}

In general, the performance of rule-based methods is not as good as that of classification models. C5.0 has better than ID3 by 6.9\% on average, probably because it takes better data discretization and segmentation mechanisms such as Gain Ratio. 
LR is implemented with discretization pre-processing which tremendously improves performance. Only the best performance of LR is shown in the table, whose discretization bin size is set as 200. 
But still, it is obvious that GBDT can achieve better results than LR, i.e., outperforms LR by 4.5\%, 2.2\%, 4.5\% and 4.2\%  for ``Basic Features'', ``Basic Feature+S2V'', ``Basic Feature+DW'' and ``Basic Features+DW+S2V'' on April 16. 


Besides F1, recalls at different thresholds are also important for real-world analysis. Such recall metric can measure the ability of the classifier to find the most suspicious fraud.
Figure \ref{fig:perf_rec} shows the recall for the top 1\% of the most suspicious cases, i.e., rec@top 1\%, over five different detection methods. 
From the results, we can see that IF performs the worst, i.e., under 10\%, which is consistent with F1. Such results are intuitive as outliers found by IF are probably not caused by fraud cases but for other reasons. Rule-based ID3 and C5.0 methods achieve much higher results, i.e., 30\% and 40\%, respectively. GBDT slightly outperforms LR and performs the best.


\begin{figure} [htbp!]
  \centering
  \includegraphics[scale=0.43]{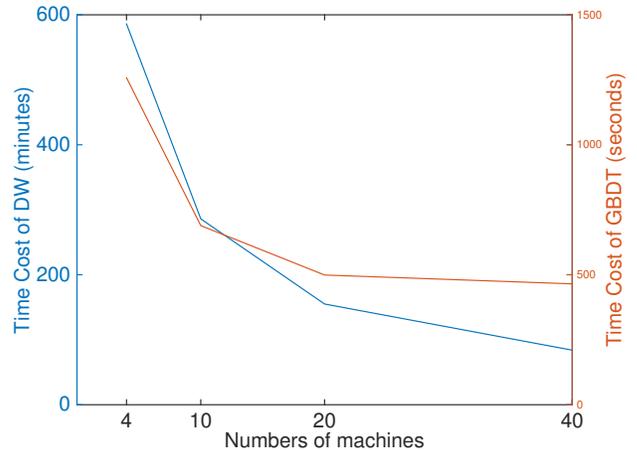}
  \caption{Time cost over the numbers of machines.}
\label{fig:perf_time}
\end{figure}

Based on the above observations, we choose DW to extract additional aggregated information. GBDT is selected as the classifier for its good performance. 
In order to decide the computing resources, we further test time cost versus the number of machines. For our reimplemented version on KunPeng, half of the machines are selected as server nodes, and the rest are used as worker nodes. 

As shown in Figure \ref{fig:perf_time}, the time cost continues to decreases as the number of the machines increases for DW. 
However, we also notice that the time cost of GBDT does not obviously halve when the number of machines increases to 40 from 20. In fact, in real-world PS environment, IO and network communication might become the bottleneck besides computation, while more machines often indicate greater communication cost due to uneven machine traffic. Moreover, in the production environment, heterogeneous tasks execute at the same time, so resource allocation is necessary to be considered. More resources requested, more waiting time may be needed for allocation. As a compromise, we finally assign 40 machines for DW and 20 machines for GBDT.


%



\subsection{Hyperparameter Sensitivity}

We further perform a hyperparameter sensitivity analysis, where Dataset 1 shown in Figure \ref{fig:time_seg} is selected for this experiment. 

\begin{figure} [htbp!]
  \centering
  \includegraphics[scale=0.46]{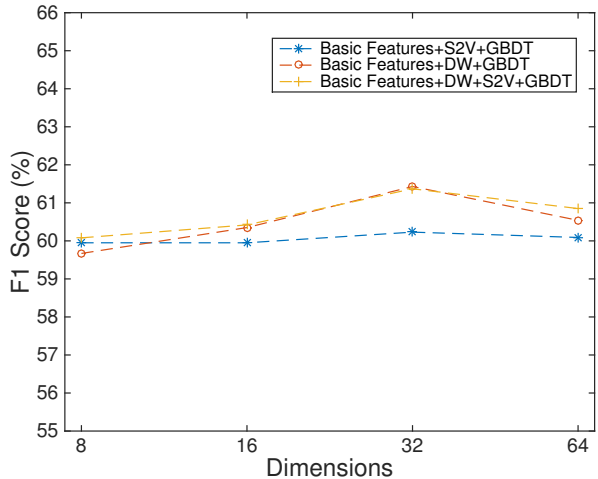}
  \caption{Performance versus the dimensions of the learned user node embeddings.}
\label{fig:perf_dim}
\end{figure}

The dimension size of the learned embeddings is an important hyperparameter, which influences the amount of topological information of the transaction network extracts. As shown in Figure \ref{fig:perf_dim}, we compare F1 score against the dimension size using different NRL methods. Obviously,  32 is the best dimension size.
We believe that the topological information of the network is not well extracted when the dimension is too small, while the results probably overfit when it is too large. 

\begin{figure} [htbp!]
  \centering
  \includegraphics[scale=0.46]{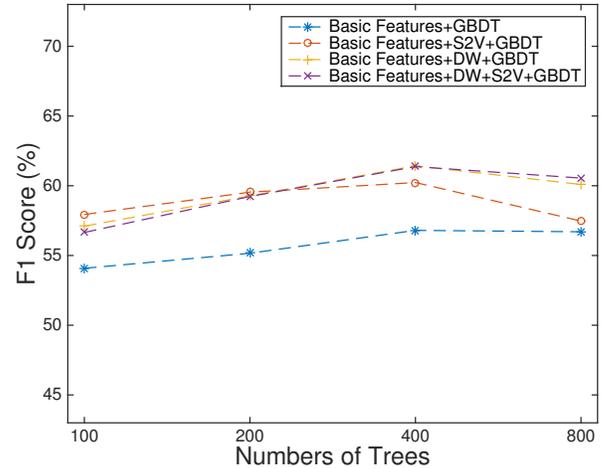}
  \caption{Performance versus the numbers of GBDT decision trees.}
\label{fig:perf_no_tree}
\end{figure}


In addition, we vary the tree size to examine the importance of tree size in GBDT.
As illustrated in Figure \ref{fig:perf_no_tree}, F1 score consistently improves as the number of trees increases to 400 and then decreases when the number of trees further increases to 800. It is intuitive that the model is not sufficiently trained when the number of trees used is too small. On the contrary, the model prone to overfitting when the number of trees is too big.

\begin{table}
\centering
\caption{Performance versus the number of node sampling.}
\vspace{6mm}
\label{table_hyper_other}
\resizebox{0.47\textwidth}{!}
{
\begin{tabular}{c|c|c|c|c} \hline
No. of Sampling&25&50&100&200 \\ \hline
F1 Score&59.67\%&60.62\%&61.43\%&61.57\% \\ \hline 
\end{tabular} }
\end{table}

Finally, we analyze the impact of the number of node sampling in DW, which controls the number of linear node sequences generated. 
Similar to the conclusion in \cite{perozzi2014deepwalk}, the performance tends to be stable as the number reaches a specific value. Table \ref{table_hyper_other} suggests that the performance tends to stabilize when the number reaches 100. Although the result is slightly better as for 200, it takes about double time to generate node sequences and learn embeddings. Besides, the depth of GBDT decision trees is also worthy of exploring, we omit it here as it can be analyzed in the similar way.

\section{Conclusion}

In this paper, we first reveal the significance of online real-time transaction fraud detection task in Ant Financial and then demonstrate our feature extraction approaches, detection models and implementation details. Extensive experiments on real-world data are conducted, showing the effectiveness and performance of our proposed TitAnt system. 
In the future, we will investigate more possibilities for system design, explore dynamic construction and modeling of a heterogeneous network, and study the interpretability of learned embeddings by NRL models.






\balance


\section{Acknowledgments}
The authors thank the anonymous reviewers for their constructive and valuable advice, and MaxCompute team for their suggestions on data storage and management, as well as Kai Xiao and Xiujing Lin's help on data preparation.

\bibliographystyle{abbrv}
\bibliography{titant}  

\end{document}